# Morphological Synthesizer for Ge’ez Language: Addressing Morphological Complexity and Resource Limitations

**Gebrearegawi Gebremariam †, Hailay Teklehaymanot *, Gebregewergs Mezgebe†**

†Axum University, Institute of Technology, Department of IT, Ethiopia
*L3S Research Center, Leibniz University Hannover,Germany
{gideygeb,gemezgebe}@aku.edu.et,teklehaymanot @l3s.de

**Abstract**

Ge’ez is an ancient Semitic language renowned for its unique alphabet. It serves as the script for numerous languages, including Tigrinya and Amharic, and played a pivotal role in Ethiopia’s cultural and religious development during the Aksumite kingdom era. Ge’ez remains significant as a liturgical language in Ethiopia and Eritrea, with much of the national identity documentation recorded in Ge’ez. These written materials are invaluable primary sources for studying Ethiopian and Eritrean philosophy, creativity, knowledge, and civilization. Ge’ez is a complex morphological structure with rich inflectional and derivational morphology, and no usable NLP has been developed and published until now due to the scarcity of annotated linguistic data, corpora, labeled datasets, and lexicons. Therefore, we proposed a rule-based Ge’ez morphological synthesis to generate surface words from root words according to the morphological structures of the language. Consequently, we proposed an automatic morphological synthesizer for Ge’ez using TLM. We used 1,102 sample verbs, representing all verb morphological structures, to test and evaluate the system. Finally, we get a performance of 97.4%. This result outperforms the baseline model, suggesting that other scholars build a comprehensive system considering morphological variations of the language.

## 1. Introduction

Language is one of the most important aspects of our lives, as it allows us to preserve information and pass it on orally or in writing from generation to generation (Allen, 1995).

Ge’ez is an ancient Semitic language with a unique alphabet ("አ፡ በ፡ ገ፡ ደ") (Adege and Mannie, 2017; Siferew, 2013). This language played a pivotal in Ethiopia and Eritrea’s cultural and religious development during the Aksumite Kingdom era. Its rich literary tradition and influence in spreading Christianity across the region are notable. Although no longer spoken colloquially after the thirteenth century, Ge’ez remains significant as a liturgical language for various religious groups. Scholars and linguists are drawn to Ge’ez for its insights into the historical evolution of Semitic languages and their connections to languages such as Hebrew, Arabic, and the modern Ethiopian and Eritrean language (Dillmann and Bezold, 2003; Desta, 2010; Abate, 2014).

Besides being the liturgical language for various religious groups in Ethiopia and Eritrea, Ge’ez remains a significant writing language for religious, historical books, and literature in the history of Ethiopia(Belcher, 2012; Scelta and Quezzaire-Belle, 2001). These written resources can be primary sources for studying Ethiopian and Eritrean philosophy, creativity, knowledge, and civilization. (Abate, 2014).

Hence, preserving the Ge’ez language becomes imperative to safeguarding Ethiopia and Eritrea’s cultural and historical heritage. As the language deeply intertwined with religious practices and literature, its preservation ensures the continuity of traditions and identities across generations. Besides, preserving the Ge’ez language is crucial for maintaining religious practices and literature traditions, honoring linguistic diversity and identity, contributing to the understanding of Semitic languages’ evolution, and fostering cultural pride and continuity across generations in Ethiopia and Eritrea (Desta, 2010).

However, research for this language has only started recently, and no usable technology has been developed and published until now for the Ge’ez because little consideration has been given to the language, even though it is that important. Due to this, Ge’ez is still a low-resource and endangered language(Eiselen and Gaustad, 2023; Haroutunian, 2022). In documenting endangered languages or reconstructing historical languages, understanding their morphological structure is essential for accurately representing and preserving the linguistic systems (Bisang et al., 2006). For morphologically rich languages such as Ge’ez, it is essential to develop a system that can generate all surface word forms from root words because this can serve as an input for many other NLP systems, including IR systems, spelling and grammar checking, text prediction, dictionary development,

POS tagging, machine translation, conversational AI, and other AI-based systems. But, it is difficult to develop AI-based systems especially for low-resourced languages such as Ge'ez, etc (Eiselen and Gaustad, 2023; Haroutunian, 2022; Gasser, 2012; Saranya, 2008; Scelta and Quezzaire-Belle, 2001; Sunil et al., 2012; Wintner, 2014).

For example, consider the search results in Table 1 to evaluate the limitation of the IR system in Ge'ez word variation.

| Queries | Verb Form | Results |
|---|---|---|
| ረጠነ/reTene/ | Perfective | 9 |
| ይረጥን/yreTn/ | Indicative | 0 |
| ይርጥን/yrTn/ | Subjective | 0 |
| ርጢን/rTin/ | Noun | 1,480 |

Table 1: Ge'ez queries and their results from the Google search engine

As shown in Table 1, the results obtained in each query are different, even though the queries are related and generated from the verb 'ረጠነ/reTene/'. In this case, the query should be given in all variants of the word forms; if not, the system will fail to retrieve the related information. However, it is inconvenient to search for all variant words (Hailay, 2013). To improve the efficiency of IR systems, it is important to create a strong relationship between the stems and their variant word forms. Thus, it is important to develop a morphological synthesizer of Ge'ez and integrate it with the IR systems to get an effective IR system.

Therefore, we proposed a rule-based Ge'ez morphological synthesizer that can play a crucial role in generating surface words from the root words according to the morphological structures of the language. This study is the first attempt to develop morphological synthesizers for the Ge'ez language, although morphological synthesizers for other languages have been developed and are available for wider usage, as stated below in the related works section. As a result, our work has made the following fundamental contributions to the scientific community:

i. We designed an algorithm based on the language's morphological rules to illustrate generating TAM and PNG features. We tried to create surface words from the lexicons. The generator uses Ge'ez Unicode alphabets without transliterating to Latin alphabets. This makes it easy to use, especially for Ge'ez learners and researchers.

ii. We prepared the first publicly available datasets for Ge'ez morphological synthesizers. Another researcher can use it.

iii. Our system gives Amharic and English meanings for the perfect verb form. Therefore, this can initiate the development of the following higher Ge'ez-Amharic, Ge'ez-Tigrinya or Ge'ez-other languages dictionary projects.

## 2. Related Works

One of the most popular research areas in NLP is the study of morphological synthesizers. Several research projects have been conducted in this area for various international languages using different approaches(Abeshu, 2013; Koskenniemi, 1983). Let us look at some related works.

ENGLEX was developed to generate and recognize English words using TLM in PC-KIMMO. It has three essential components, including a set of phonological (or orthographic) rules, lexicons (stems and affixes), and grammar components of the word. The generator accepts lexical forms such as **spy** + **s** as input and returns the surface word spies. The online source code is available here[1].

Jabalín was developed for both analyzing and generating Arabic verb forms using Python. They created a lexicon of 15,453 entries. This was designed using a rule-based approach called root-pattern morphology. The morphological generator accepts verb lemmas to produce inflected word forms and achieved an accuracy of 99.52% for correct words (González Martínez et al., 2013).

Using a paradigm-based approach, the Morphological Analyzer and Synthesizer for Malayalam Verbs was also developed by (Saranya, 2008). This helps in creating an English-Malayalam machine translation system.

Pymorphy2 was developed for the morphological analysis and generation of Russian and Ukrainian languages (Korobov, 2015). The system used large and efficiently encoded lexicons built from Open-Corpora and LanguageTool data. A set of linguistically motivated rules was developed to enable morphological analysis and the generation of out-of-vocabulary words observed in real-world documents.

TelMore was developed by (Ganapathiraju and Levin, 2006) to handle the morphological generation of nouns and verbs in Telugu. The prototype was designed based on finite-state automata. TelMore accepts the infinitive form for the verb types and generates the present, past, and future tenses, affirmative, negative, imperative, and prohibitive forms for all genders and numbers. In addition, (Dokkara et al., 2017) also developed a morphological generator for this language. Its computational model was developed based on finite-state techniques. The system was evaluated for a total

[1] http://downloads.sil.org/legacy/pc-kimmo/engl20b5.zip

of 503 verbs. Of these verbs, 418 words were correct, and 85 words were incorrect.

(Goyal and Lehal, 2008) developed the morphological analyzer and generator for Hindi using the paradigm approach. This system has been developed as part of the machine translation system from Hindi to Punjabi. (Gasser, 2012) developed a system that generates words for Amharic, Oromo, and Tigrinya words from the given root and affixes. This has been developed based on the concept of finite-state technology. The system produced 96% accurate results (Gasser, 2012).

A morphological synthesizer for Amharic was developed by (Lisanu, 2002) using combinations of rule-based and artificial neural network approaches. However, his study was limited to Amharic perfect verb forms. Some of the generated word forms could be more meaningful. Also, this model used a transliteration of the Amharic script into Latin before any synthesis was done. The system does not allow generation for other roots that are not registered in its database. On the other hand, words are generated as output by giving the root and suffix as inputs. This may limit the number of words the model can produce compared to the words developed by the language experts. (Lisanu, 2002).

(Abeshu, 2013) developed an automatic morphological synthesizer for Afan Oromoo using a combination of CV-based and TLM-based approaches and achieved a performance of 96.28 % for verbs and 97.46% for nouns. The study indicated that developing a full-fledged automatic synthesizer for Afan Oromoo using rule-based approaches can yield an outstanding result. And it is easy to extend the system to other parts of speech with minimal effort.

The morphological synthesizers reviewed overhead are specific to their corresponding language and cannot handle Ge'ez's morphological characteristics because Ge'ez differs from these languages. To our knowledge, no research has been conducted to develop an automatic morphological generator for the Ge'ez language. Thus, we planned to create a morphological synthesizer model that can generate the derivational and inflectional morphology of Ge'ez language verbs.

## 3. Ge'ez Morphology

Ge'ez language has a complex morphological structure because a single word can appear in many different forms and convey different meanings by adding affixes or changing the phonological patterns of the word (Adege and Mannie, 2017). In particular, verbs have a more complex structure than other POSs in Ge'ez. Thus, Ge'ez verbs are categorized into six principal classes in their forms labeled as perfective, indicative, infinitive, subjunctive, jussive, and gerundive verb forms. Each verb form has five stem classes, and each verb stem will inflect by adding affixes to create different word forms (Desta, 2010). Generally, there are three phases to creating variant word forms in Ge'ez, as defined in (Dillmann and Bezold, 2003). These are given below, as depicted in Figure 1:

Phase I: Stem formation

Phase II:TAM formation

Phase III: PNG formation

In Phase I, the declaration of word forms using the Tense-Mood as rows and the five stems as columns is done.

In Phase II, each surface verb form obtained from Phase I is further declared using the ten subjective pronouns by appending the subject marker suffix.

In Phase III, declarations of the word forms using the ten Object Maker Suffixes for each of the words obtained in Phase II will occur.

So, two rules for suffixing verbs govern the concatenation process of morphemes to produce the surface verb forms:

- Stem + subject-marker suffix = surface word (only with SMS)
- Stem + subject-marker suffix + object-indicator suffix = surface word (with both SMS and OMS)

Hence, we can have two verb forms, one with the only direct subject marker and the other with both subject marker and object marker suffixes, as indicated below:

**ቀተል** (stem) + **ክሙ** (subject marker suffix) = **ቀተልክሙ** - you killed. (Surface Form).

**ቀተል** (stem) + **ክሙ** (object marker suffix) = **ቀተለክሙ** - he killed you (Surface Form).

**ቀተል** (stem) + **ክሙ** (SMS) + **ኒ** (OMS) = **ቀተልክሙኒ** - you killed me (Surface Form).

In this case, the subject marker suffix /-**ክሙ**/ points out that the subject is "you (2 ppm)," whereas the object marker /-**ኒ**/ indicates the object "me.". Hence, the verb /**ቀተልክሙኒ**/ indicates both the subject and the object of the verb. Hence, a single verb can be a sentence in Ge'ez because it has both subject and object indicator suffixes.

## 4. Methodology of the study

We have reviewed several books, research reports, journals, articles, and user manuals to grasp the morphological structure of Ge'ez verbs and to know the different techniques for designing morphological synthesizers. In addition, continuous discussions were conducted with Ge'ez experts to better understand the morphological structure of the language better and to get valuable ideas for the study.

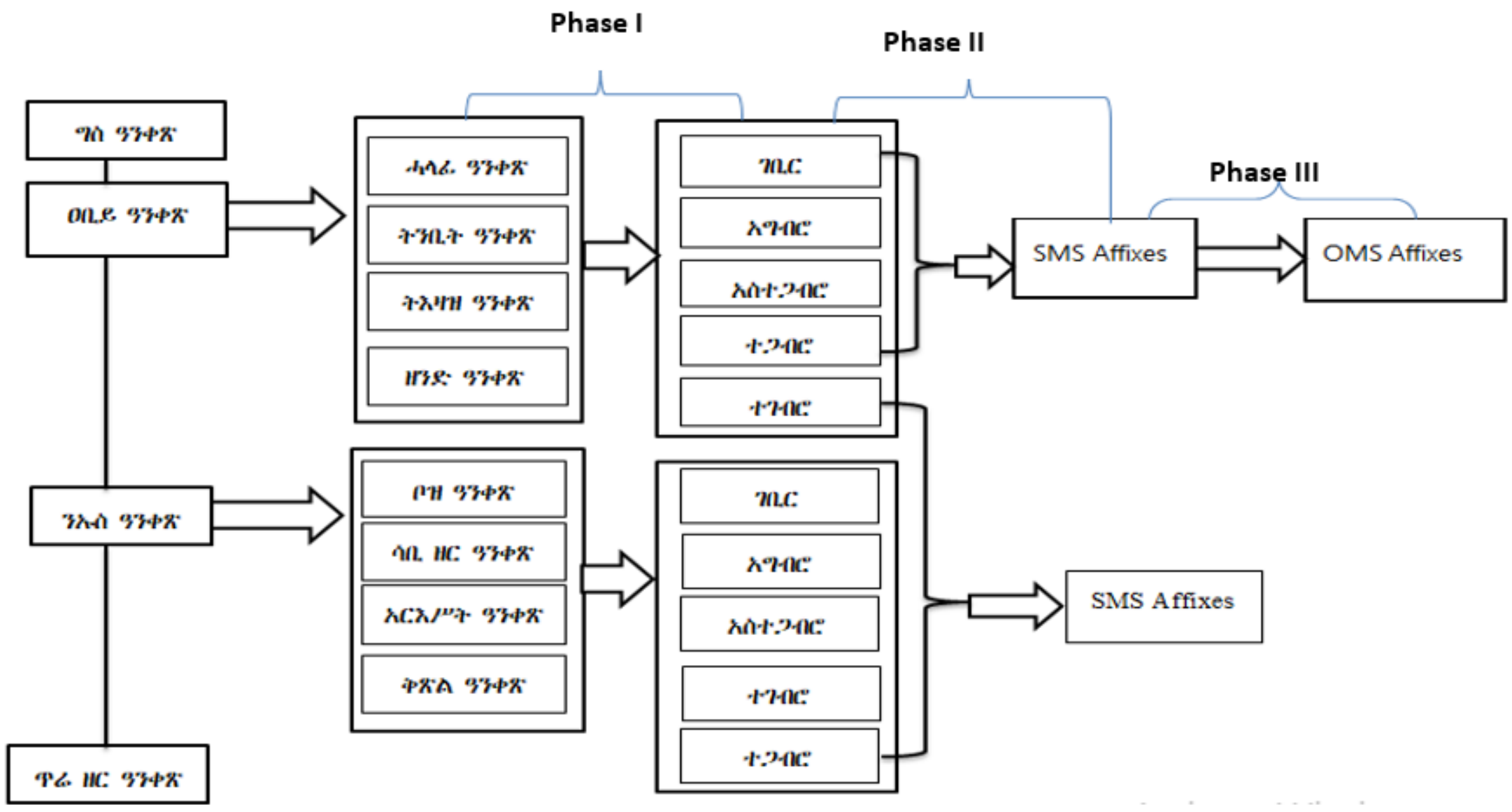

Figure 1: Phases of Ge'ez morphological word formation

### 4.1. Data Collection

Manually annotated data in lexicons helps test the morphological synthesizer. Since machine-readable dictionaries and word lists or an online corpus for Ge'ez were not available, the work of compiling the lexicons was started from scratch. Hence, we have compiled sample representative verbs that characterize all variations of verbs for testing and evaluating the systems's performance by consulting experts of the language. These verbs are collected from different books, like the Holy Bible, **መጽሐፈ ግስ**/Ge'ez Grammar Book/, and from Isanate siem (**ልሳናተ ሴም**) (Zeradawit, 2017). Therefore, the language lexicon prepared for this study consists of 1102 regular and irregular verbs. The affixes that can be concatenated with the verbs are also compiled into the lexicons.

### 4.2. Design

As defined by (Pulman et al., 1988), it is mandatory to consider at least the following basic design requirements to develop a morphological synthesizer of a language:

**1. Lexicons:**

Lexicon describes the list of all lemmas and all their forms. It is the heart of any natural language processing system, even though the format differs according to their needs. Consequently, the lexicons required for our study include stems, affixes, and Ge'ez alphabets. Let us see each of these lexicons in detail. i. **Stems:** In our study, the stem inputs are infinitive verb forms like **ቀቲል**/to kill/, **ሐዊር**/to walk/, **ሰጊድ**/to Prostrate/, **ፈቂድ**/to allow/, **ሐዩው**/to salivate/, etc. From these lexical inputs, the system generates inflected words for all genders and numbers by combining them with the corresponding affixes according to the set of rules of the language. The reason why we want to use the infinitive verb form as input instead of the root word/**ጥሬ ዘር**/ is to remove the ambiguity that may be created when the prototype distinguishes the input's verb category.

ii. **Affixes:** As defined by (Abebe, 2010), the affixes carry different types of syntactic and semantic information, helping to construct various words. Affixes combine with the word stems to generate various words based on the set of rules. Here, Verbal-Stem-Marker Prefixes and Person-Marker Prefixes are combined first with the input stem to generate various word stems (Abebe, 2010). Then, SMS and OMS suffixes follow in sequence. For example, consider the formation of **ይቀትለከ** /He will kill you/ using TLM in Table 2.

As indicated in Table 2, for every stem to combine with affixes, an analyzer should investigate the type of stem and the affixes that can concatenate properly to create valid surface words. Hence, a set of rules was established to handle such requirements.

iii. Ge'ez Alphabets: As described by (Koskenniemi, 1983), both the lexical and surface-level words in the two-level model are strings extracted from the language alphabets. The lexical-level strings may contain some characters that may not occur on the surface-level strings. Accordingly, Ge'ez words are constructed by the meaningful concatenation of Ge'ez alphabets. The alphabets in the Ge'ez language include all the characters starting from **ሀ**/he/ to **ፈ**/fe/ and the four other complex-compound alphabets. All the alternations of characters in the lexical strings during surface word formations are retrieved from these alphabets. Implementing these alterations is handled based on the rules in the system prototype. The two-level rules are used here to specify the permis-sible differences between lexical and surface word representations.

2. **Morphotactics**:

Morphotactics is the model or rule of morpheme ordering that explains which classes of morphemes can follow other courses of morphemes inside a word. Ge‘ez verbs have their own rules for ordering the morphemes. The order of morphemes in the word formation of Ge’ez verbs is as follows:

[Prefix] + [Prefix Circumfixes] + [stem] + [Suffix Circumfixes] + [SMS] + [OMS]

3. **Orthographic Rules:**

Orthographic rules are the spelling rules that are used to model the changes that occur in a word when two morphemes combine. Therefore, a set of rules is essential in mapping input stems to surface word forms. These rules are designed based on the morphological nature of Ge’ez for each sequence of the word formation process. Ge’ez has its own spelling rules when morphemes are concatenated with each other. For example **ቀተለ**/qetele/+-**ኩ**/ku/: **ቀተልኩ**/qetelku/ (here, **ለ**/le/ is changed to /l/ when {**ቀተለ**/qetele/} is added to the SMS {**ኩ**/ku/}).

By taking the above design requirements into account, we designed the general flow chart of the system as shown in Figure 2: As we see in the flow chart in Figure 2, the design of morphological synthesizer has the following components:

**A.Stem Classifier:** identifies the verb category of the stem. The classification is undertaken based on the number of heads and troops of verbs. This component also checks whether the verb stem is regular or not. Here, if the input verb contains one of the guttural alphabets (namely **ሀ**/he/, **ሐ**/He/, **ኀ**/H/, **አ**/a/ and **ዐ**/A/ either at their beginning or middle positions) or semi-vowel alphabets (namely **የ**/ye/ and **ወ**/we/) at any positions of the verb, it is irregular, else it is regular verb.

**B.Stems Formation:** This sub-component generates the various derived stems for the lexical input.

**C.Signature Builder:** lists the set of suffixes valid for each generated stem because every created stem has specific corresponding affixes to the stem during valid surface word formation. To establish a valid concatenation of the stems with affixes, a pattern matching mechanism is used, which is based on the notion of matching the stems with their valid affixes. For example, the word ‘**ይቀትል**’/yqetl/ has a valid affix ‘**ዎ**’/wo/ to create a valid word form. But, this word cannot be combined with the affix ‘**ክዎ**’/kwo/ because the combination of the word and the affix cannot create valid word forms.

**D. Boundary Change Handler:** This sub-component addresses the boundary patterns occurring during the concatenation of stems and affixes based on the rules laid down on the knowledge base. These changes may be specific to every morpheme concatenation, even if these morphemes are in the same manner. Assimilation effects are occurring mostly on the boundary of the morphemes when the suffixes **ከ** /ke/, **ኩ** /ku/, **ኪ** /ki/, **ክን** /kn/ or **ክሙ** /kmu/ are added to the end of a verb that ends with either of the glottal alphabets, namely **ቀ** /QE/, **ከ** /ke/, or **ገ** /Ge/ (Lambdin, 1978). For example, observe the concatenation of the morphemes **ሐደገ** with **ክሙ**:

**ሐደገ** + **ክሙ** –> **ሐደግሙ** (the character **ገ** in **ሐደገ** changes to **ግ** and the character **ክ** is omitted from the morpheme **ክሙ**)

**E. Synthesizer:** This sub-component generates all possible surface word forms by concatenating the stem with the selected list of affixes using the TLM method of word generation. For example, consider the following Ge’ez word generation by TLM from Table 2:

| | | | | | | |
|---|---|---|---|---|---|---|
| **Lexical Level** | ይ | ቀ | ት | ል | + | ክ |
| **Surface Level** | ይ | ቀ | ት | ለ | 0 | ክ |

Table 2: Generation of surface words using TLM

The rows in Table 2 depict the two-level mappings carried out during the word formation process.

**F. Surface Level:** Lastly, the outputs of the synthesizer are produced.

Below is our concise algorithm for producing word forms based on input lexicons:

..........................................................

1. Start
2. Input infinitive verb stem (verb stem)
3. Classify verb regularity using classifyVerbRegularity(verbstem)
4. If regular:
4.1 For each stem in generateStems(verb stem):
4.1.1 Select affixes with selectAffixes(stem)
4.1.2 Apply boundary changes with applyBoundaryChanges(stem)
4.1.3 Concatenate changed stems with affixes
4.1.4 Print output words
5. Else (if irregular):
5.1 For each stem in generateStems (verbstem):
5.1.1 Select affixes with selectAffixes(stem)
5.1.2 Apply boundary changes with applyBoundaryChanges(stem)
5.1.3 Concatenate changed stems with affixes
5.1.4 Print output words
6. End

# 5. Experimentation and Evaluation

## 5.1. Developmental Approach

Several approaches could have been applied to developing morphological generation systems for different languages. As discussed by (Kazakov and Manandhar, 2001), these approaches can be categorized as rule-based and corpus-based approaches. This study applied the rule-based approach called the Two-Level Model (TLM) of morphology to develop the prototype. TLM is used to handle the phonological and morphophonemic processes (including assimilation changes) involved in word formation (Gasser, 2011; Koskenniemi, 1984). Principally, we selected the TLM approach to map lexical entries to surface verb forms. We used the rule-based approach to develop the morphological synthesizer of the language because this approach has a faster development process with better accuracy, is more straightforward to twist, and is more accessible for formulating rules according to the language rules (Beesley and Karttunen; Shaalan et al., 2007). Moreover, the rule-based approach is practical for languages with fewer resources, such as Ge'ez, which suffers from the availability of corpora and the scarcity of data(Shaalan et al., 2010). Hence, we preferred the rule-based approach, in which a particular word is given as an input to the morphological synthesizer, and if that corresponding morpheme or root word is valid, then the system will produce surface word forms.

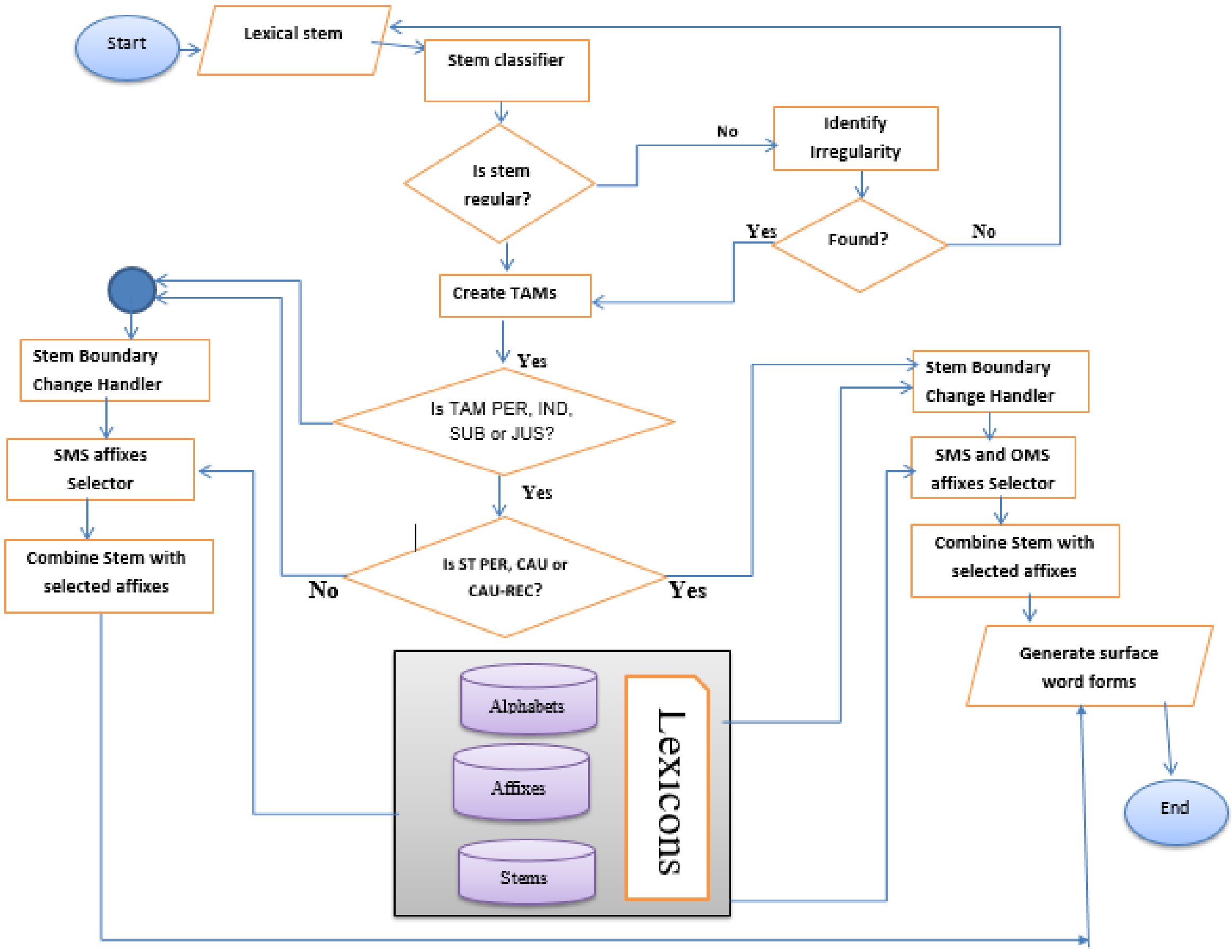

Figure 2: Flow Chart of Ge'ez morphological synthesizer

### 5.2. Testing Procedures

Systematic evaluation of the system is complex since no collected Ge'ez words are currently available for this purpose. So, to test the effectiveness of the system developed, we used the collected sample verbs. The testing procedures are as follows:

**1.**During the initial phase, we evaluated the system by inputting a test stem extracted from sample verbs in the lexicon, generating words, and comparing them with their expected word forms. This evaluation was conducted iteratively throughout the development of the morphological synthesizer to enhance its performance. Any errors identified during this testing, primarily related to missing rules, were rectified accordingly. Subsequent iterations of this test were conducted until satisfactory results were achieved.

**2.**Then, the finalized system's functionality was tested by entering sample verbs (including those with glottal or semivowel alphabets at different positions) selected by linguists.

### 5.3. Evaluation Procedures

Finally, we evaluated by taking regular and irregular verbs from the selected sample verbs. To evaluate the system, we used two options:

**1.Manual Evaluation** Using the error-counting approach, language experts manually evaluated the generated words to assess their accuracy and quality. The system accuracy is then calculated as the number of correctly generated words divided by the total number of words generated by the system multiplied by 100%.

**2.Automatic Evaluation** We evaluate system performance using predefined criteria and metrics without human intervention, a method akin to that described by (González Martínez et al., 2013). Subsequently, the accuracy attained from each experiment is calculated using the following formula:

$$\textbf{Accuracy} = \frac{Correctlygeneratedwords}{totalgeneratedwords} * 100 \quad (1)$$

## 6. Experimental Results

The accuracy assessment of the developed system involved inputting sample datasets. 7,577 words were generated from regular verbs, and 19,290 words were generated from irregular verbs. Out of these, 668 errors were identified (8 from regular verbs and 661 from irregular verbs). The accuracy rates were: 99.6% for regular verbs and 96.6% for irregular verbs. Resulting in an overall average accuracy of 97.4%. This result Surpasses the baseline(Abeshu, 2013). The percentage of words with errors was 2.6%. This promising outcome supports further research on the language. The experimental results are found and referred in the Appendix section.

## 7. Discussion

The system consistently produces accurate words, albeit with occasional errors. As Appendix I details, irregular verbs perform less than regular verbs, primarily due to their inherent flexibility in word-formation processes. The predominance of irregular verbs in the evaluation dataset contributes to the observed decrease in accuracy. If a more significant proportion of regular verbs were included in the evaluation, the accuracy would be expected to surpass 97.4%,given the higher accuracy rate of 99.6% observed for words generated from regular verbs.

### 7.1. Factors Leading to High Performance.

Despite encountering some errors, the synthesizer demonstrates remarkably high performance. This achievement can be attributed to several factors:

**1.Creating correct stems** Correctly generated stems generate correct surface words if the boundary changes happening during stem and affix concatenations are handled correctly. If the root stems are developed perfectly, then the words generated from these stems are correct. Hence, the performance achieved is high because most of the stems caused are right, and the boundary changes are handled correctly.

**2. Handling of rules when morphemes are concatenated with each other** Correct words are generated when stems and affixes are concatenated properly. For this reason, the selection of affixes for the given stem was handled properly. Therefore, handling the set of rules for word formation properly will generate valid words.

**3. Handling rules for irregular word formation** Ge'ez language has many irregular verbs. Irregular verbs are those that have a slight change in their morphological structure when compared to regular verbs. This is mostly happening due to the existence of one of the guttural alphabets, namely **ሀ**/he/, **ሐ**/He/, **ኀ**/H/, **አ**/a/ and **ዐ**/A/ either at their beginning or middle positions, or the existence of the semi-vowel alphabets, namely **የ**/ye/ and **ወ**/we/ at either position of the verbs. Irregular verbs have various rules to generate the correct word forms. These rules have slight differences from these regular word formation rules. Handling these rules of word formation gives you better accuracy. Accordingly, we have tried to handle the word formation rules as much as possible.

### 7.2. Error Analysis

Certain words are generated incorrectly. These errors can be attributed to the following factors:

**1. Errors caused due to exceptional characters existing in the verb** Some verbs have special characteristics, even though these verbs seem to have the same form as the head verb. For example, the system was designed to handle the verbs that end with the characters **ቀ**/qe/, **ከ**/ke/, and **ገ**/ge/ because it is assumed that these verbs have the same morphological characteristics as other verbs. However, this may not always be true if we consider the morphological structure of the verbs **ሠረቀ**/šereqe/, **ሐደገ**/Hedege/, and **ለሐቀ**/leHeqe/. These verbs have differences due to the existence of guttural or semi-vowel characters, or both as shown in Table 3.

| Verbs | Differences observed | | | |
|---|---|---|---|---|
| | Indicative | Subjective | Jussive | Infinitive |
| ሠረቀ/sereke/ | ይሠርቅ/yserk/ | ይሥርቅ/ysrk/ | ይሥርቅ/ysrq/ | ሠሪቅ/seriq/ |
| ሐደገ/Hedege/ | የሐድግ/yeHedg/ | ይሕድግ/yHdg/ | ይሕድግ/yHdq/ | ሐዲግ/Hedig/ |
| ለሐቀ/leHeqe/ | ይልሕቅ/ylHq/ | ይልሐቅ/ylHaq/ | ይልሐቅ/ylHaq/ | ልሒቅ/lHiq/ |

Table 3: Errors caused by exceptional characters

As we see in table 3, the letters written in red color in the words make a difference in each word formation process even though these words are categorized in the same verb category.

**2. Errors generated during concatenation of exceptional words with affixes**

Some of the generated words seem to be correct both grammatically and semantically, but they are not correct words. For example, when the morphemes ከረም/kerem/ and ነ/ne/ are concatenated, they produce the word ከረምነ/keremne/ which is the correct word. In the same way, when the morphemes አመን/amen/ + ነ/ne/, it gives አመንነ/ amenne/. However, አመንነ/ amenne/ is not the correct word. The correct word is አመነ/amene/. So, these words have different forms even though they belong to the same POS and number.

**3.Errors caused due to morphological richness and varied nature of the language**

This type of error occurs when testing with verbs that seems to have the same structure as other verbs in nature. But their actual output shows different word forms. For example, when we take the verbs ወለደ/welede/ and ወቀሰ/weqese/, we assume that these verbs have the same structure during the design of the prototype. But these verbs have differences in their actual word formation structure.

**4. Errors caused by missing some rules** The formation of the different word forms has a set of rules. Missing any of these rules generates invalid word forms. The incorrect words in Table 4 are generated because some rules and their correct forms are missing.

| መራሒ ግስ (pronoun) | Incorrectly generated words | Correct words |
|---|---|---|
| ውእቱ (He) | ተከብበ/tekebbe/ | ተከበ/tekebe/ |
| ይእቲ (She) | ተከብበት/tekebbet/ | ተከበት/tekebet/ |
| ውእቶሙ (They-male) | ተከብቡ/tekebbu/ | ተከቡ/tekebu/ |
| ውእቶን(They-female) | ተከብባ/tekebba/ | ተከባ/tekeba/ |

Table 4: Errors caused due to missing rules

## 8. Conclusion and Future Work

The study opted for the rule-based TLM approach for developing an automatic morphological synthesizer due to its simplicity, suitability, and effectiveness, especially for languages with limited corpora availability. A set of rules was meticulously designed based on expert knowledge of the language's morphological structure, forming the foundation for algorithm development from scratch to handle word formation processes. Despite the thoroughness of the morphological synthesis rules, some inaccuracies persisted in word generation, mainly stemming from the formation of invalid stems, notably for irregular verbs containing guttural and semi-vowel alphabets. Nevertheless, the prototype synthesizer exhibited promising performance, with an overall accuracy of 97.4%, indicating encouraging prospects for further research in Ge'ez linguistics. Feedback from linguists involved in the system evaluation underscored the importance of developing a comprehensive system version to enhance Ge'ez's usage and preservation within society. Recommendations were made for future researchers to address and rectify errors limiting the study's performance and to advance toward a fully functional system. Challenges encountered during the study included: A lack of Ge'ez linguistic experts. Absence of standardized references and dictionaries. Scarcity of compiled Ge'ez language lexicons. Furthermore, the complexity and agglutinative nature of Ge'ez morphology posed additional hurdles, contributing to its extensive vocabulary.

## List of Acronyms

EOTC…………Ethiopian Orthodox Tewahido Church
TLM…………Two-Level Morphology
NLP…………..Natural Language Processing
PNGs………...Persons, Numbers and Genders
POS………….Parts Of Speech
TAM…………Tense-Aspect-Mood
SMS………….Subject Marker Suffixes
OMS…………Object Marker Suffixes
IR…………….Information Retrieval
CV……………Consonant-Vowel
PER…………..Perfective
IND…………..Indicative
SUB………….Subjective
JUS…………..Jussive
ST……………Stem Type
CAU………….Causative
CAU-REC… ...Causative-Reciprocal
XML………….Extensible Markup Language

## References

Yitayal Abate. 2014. Morphological analysis of ge'ez verbs using memory based learning.

A. Abebe. 2010. Automatic morphological synthesizer for afaan oromoo. A thesis Submitted to School of Graduate Studies of addis ababa University in Partial fulfillment for degree masters of Science in Computer Science.

Abebe Abeshu. 2013. Analysis of rule based approach for afan oromo automatic morphological synthesizer. *Science, Technology and Arts Research Journal*, 2(4):94–97.

Abebe Belay Adege and Yibeltal Chanie Mannie. 2017. *Designing a Stemmer for Ge'ez Text Using Rule based Approach*. LAP LAMBERT Academic Publishing.

James Allen. 1995. *Natural language understanding*. Benjamin-Cummings Publishing Co., Inc.

Kenneth R Beesley and Lauri Karttunen. Finite-state morphology: Xerox tools and techniques. *CSLI, Stanford*, pages 359–375.

Wendy Laura Belcher. 2012. *Abyssinia's Samuel Johnson: Ethiopian Thought in the Making of an English Author*. OUP USA.

Walter Bisang, Hans Henrich Hock, Werner Winter, Jost Gippert, Nikolaus P Himmelmann, and Ulrike Mosel. 2006. *Essentials of language documentation*. Mouton de Gruyter.

Berihu Weldegiorgis Desta. 2010. *Design and Implementation of Automatic Morphological Analyzer for Ge'ez Verbs*. Ph.D. thesis, Addis Ababa University.

August Dillmann and Carl Bezold. 2003. *Ethiopic grammar*. Wipf and Stock Publishers.

Sasi Raja Sekhar Dokkara, Suresh Varma Penumathsa, and Somayajulu G Sripada. 2017. Verb morphological generator for telugu. *Indian Journal of Science and Technology*, 10:13.

Roald Eiselen and Tanja Gaustad. 2023. Deep learning and low-resource languages: How much data is enough? a case study of three linguistically distinct south african languages. In *Proceedings of the Fourth workshop on Resources for African Indigenous Languages (RAIL 2023)*, pages 42–53.

Fitsum Gaim, Wonsuk Yang, and Jong C Park. 2022. Geezswitch: Language identification in typologically related low-resourced east african languages. In *Proceedings of the Thirteenth Language Resources and Evaluation Conference*, pages 6578–6584.

Madhavi Ganapathiraju and Lori Levin. 2006. Telmore: Morphological generator for telugu nouns and verbs. In *Proceedings of the Second International Conference on Digital Libraries*.

Michael Gasser. 2011. Hornmorpho: a system for morphological processing of amharic, oromo, and tigrinya. In *Conference on Human Language Technology for Development, Alexandria, Egypt*, pages 94–99.

Michael Gasser. 2012. Hornmorpho 2.5 user's guide. *Indiana University, Indiana*.

Alicia González Martínez, Susana López Hervás, Doaa Samy, Carlos G Arques, and Antonio Moreno Sandoval. 2013. Jabalín: a comprehensive computational model of modern standard arabic verbal morphology based on traditional arabic prosody. In *Systems and Frameworks for Computational Morphology: Third International Workshop, SFCM 2013, Berlin, Germany, September 6, 2013 Proceedings 3*, pages 35–52. Springer.

Alicia González Martínez, Susana López Hervás, Doaa Samy, Carlos G. Arques, and Antonio Moreno Sandoval. 2013. Jabalín: A comprehensive computational model of modern standard arabic verbal morphology based on traditional arabic prosody. In *Systems and Frameworks for Computational Morphology*, pages 35–52, Berlin, Heidelberg. Springer Berlin Heidelberg.

Vishal Goyal and Gurpreet Singh Lehal. 2008. Hindi morphological analyzer and generator. In *2008 First International Conference on Emerging Trends in Engineering and Technology*, pages 1156–1159. IEEE.

B. Hailay. 2013. Design and development of tigrigna search engine. A thesis Submitted to School of Graduate Studies of addis ababa University in Partial fulfillment for the Degree of Master of Science in Computer Science.

Levon Haroutunian. 2022. Ethical considerations for low-resourced machine translation. In *Proceedings of the 60th Annual Meeting of the Association for Computational Linguistics: Student Research Workshop*, pages 44–54.

Dimitar Kazakov and Suresh Manandhar. 2001. Unsupervised learning of word segmentation rules with genetic algorithms and inductive logic programming. *Machine Learning*, 43:121–162.

Mikhail Korobov. 2015. Morphological analyzer and generator for russian and ukrainian languages. In *Analysis of Images, Social Networks and Texts: 4th International Conference, AIST 2015, Yekaterinburg, Russia, April 9–11, 2015, Revised Selected Papers 4*, pages 320–332. Springer.

Kimmo Koskenniemi. 1983. *Two-level morphology: A general computational model for word-form recognition and production*. University of Helsinki. Department of General Linguistics.

Kimmo Koskenniemi. 1984. A general computational model for word-form recognition and production. In *10th International Conference on Computational Linguistics and 22nd Annual Meeting of the Association for Computational Linguistics*. The Association for Computational Linguistics.

Thomas O. Lambdin. 1978. *Introduction to Classical Ethiopic (Ge'ez)*. Harvard Semitic Studies - HSS 24.

K Lisanu. 2002. *Design and development of automatic morphological synthesizer for Amharic perfective verb forms*. Ph.D. thesis, Master's thesis, school of Information Studies for Africa, Addis Ababa.

Stephen G Pulman, Graham J RUSSELL, Graeme D Ritchie, and Alan W Black. 1988. Computational morphology of english.

SK Saranya. 2008. Morphological analyzer for malayalam verbs. *Unpublished M. Tech Thesis, Amrita School of Engineering, Coimbatore*.

Gabriella F Scelta and Pilar Quezzaire-Belle. 2001. The comparative origin and usage of the ge'ez writing system of ethiopia. *Unpublished manuscript, Boston University, Boston. Retrieved July*, 25:2009.

Khaled Shaalan, Azza Abdel Monem, and Ahmed Rafea. 2007. Arabic morphological generation from interlingua: A rule-based approach. In *Intelligent Information Processing III: IFIP TC12 International Conference on Intelligent Information Processing (IIP 2006), September 20–23, Adelaide, Australia 3*, pages 441–451. Springer.

Khaled Shaalan et al. 2010. Rule-based approach in arabic natural language processing. *The International Journal on Information and Communication Technologies (IJICT)*, 3(3):11–19.

Muluken Andualem Siferew. 2013. *Comparative classification of Ge'ez verbs in the three traditional schools of the Ethiopian Orthodox Church*, volume 17 of *Semitica et Semitohamitica Berolinensia*. Shaker Verlag, Aachen.

R Sunil, Nimtha Manohar, V Jayan, and KG Sulochana. 2012. Morphological analysis and synthesis of verbs in malayalam. *ICTAM-2012*.

Shuly Wintner. 2014. Morphological processing of semitic languages. In *Natural language processing of Semitic languages*, pages 43–66. Springer.

A. Zeradawit. 2017. **ልሳናተ ሴም**, 1st edition. **ትንሳኤ ማተምያ ድርጅት**, Addis Ababa, Ethiopia.

# Appendix I

| No. | Verb Input | Verb Form | Number of words Generated | Number of correctly Generated words | Number of wrongly Generated words | Accuracy |
|---|---|---|---|---|---|---|
| 1. | ፈቀደ/feqede/ | Regular | 1269 | 1269 | 0 | 100% |
| 2. | አመነ/amene/ | Irregular | 590 | 563 | 27 | 95.4% |
| 3. | ሠረቀ/šereqe/ | Irregular | 1262 | 1262 | 0 | 100% |
| 4. | ከደነ/kedene/ | Regular | 1260 | 1233 | 27 | 97.9% |
| 5. | ሰበከ/sebeke/ | Irregular | 1262 | 1262 | 0 | 100% |
| 6. | ሐደገ/Hedege/ | Irregular | 1262 | 1162 | 100 | 92.0% |
| 7. | መሐለ/meHele/ | Irregular | 580 | 547 | 33 | 94.3% |
| 8. | ቀነየ/qeneye/ | Irregular | 580 | 580 | 0 | 100% |
| 9. | አበየ/abeye/ | Irregular | 580 | 490 | 90 | 84.4% |
| 10. | ጠወየ/Teweye/ | Irregular | 580 | 570 | 10 | 98.2% |
| 11. | ጠዐመ/TeAme/ | Irregular | 1162 | 1162 | 0 | 100% |
| 12. | ሐጸየ/Hetseye/ | Irregular | 580 | 490 | 90 | 84.4% |
| 13. | ዘበጠ/zebeTe/ | Regular | 1262 | 1262 | 0 | 100% |
| 14. | ሐመመ/Hememe/ | Irregular | 580 | 580 | 0 | 100% |
| 15. | ወለደ/welede/ | Irregular | 1262 | 1262 | 0 | 100% |
| 16. | ሐረደ/Herede/ | Irregular | 580 | 580 | 0 | 100% |
| 17. | ሐለየ/Heleye/ | Irregular | 580 | 490 | 90 | 84.5% |
| 18. | ፈደየ/fedeye/ | Irregular | 580 | 580 | 0 | 100% |
| 19. | ከወወ/kewewe/ | Irregular | 580 | 576 | 4 | 99.3% |
| 20. | ተለወ/telewe/ | Irregular | 580 | 580 | 0 | 100% |
| 21. | ከበበ/kebebe/ | Regular | 1262 | 1258 | 4 | 99.7% |
| 22. | ሐተተ/Hetete/ | Irregular | 580 | 576 | 4 | 99.3% |
| 23. | ወጠነ/weTene/ | Irregular | 580 | 551 | 29 | 95% |
| 24. | ረወየ/reweye/ | Irregular | 584 | 584 | 0 | 100% |
| 25. | ለወሰ/lewese/ | Irregular | 1262 | 1262 | 0 | 100% |
| 26. | ጸደለ/tsedele/ | Regular | 1262 | 1262 | 0 | 100% |
| 27. | ዘረወ/zerewe/ | Irregular | 580 | 580 | 0 | 100% |
| 28. | ገደፈ/gedefe/ | Regular | 1262 | 1262 | 0 | 100% |
| 29. | ገረመ/gereme/ | Regular | 1262 | 1262 | 0 | 100% |
| 30. | ወቀሰ/weqese/ | Irregular | 1262 | 1082 | 180 | 85.7% |
| **Total** | | | **26,867** | **26,179** | **688** | |
| | | | | | **Average Accuracy** | **97.4%** |

Results obtained by the experimentation of the system prototype

| Prefixes | | Suffixes | | | Circumfixes |
|---|---|---|---|---|---|
| ኢ | ያስተ | ኩ | እ | ሆሙ | እ-እ |
| አ | አስተ | ነ | ኢ | ዮሙ | ን-እ |
| ያ | እ | ከ | ትየ | ዋ | ት-እ |
| ይ | ን | ኪ | ትነ | ያ | ት-ኡ |
| ት | እት | ክሙ | ን | ዎን | ት-ኢ |
| ታ | ንት | ክን | ከ | ሆን | ት-አ |
| ይት | ተ | አ | ሃ | ዮን | ይ-እ |
| ትት | ና | ኡ | ሁ | ኒ | ይ-ኡ |
| ታስተ | የ | አት | ዎ | ኮ | ይ-አ |
| ነ | ዘ | ኣ | ዮ | ቱ | |
| አስ | ለ | የ | ሙ | ዎሙ | |
| ናስተ | | ኦ | | | |

Some of the Identified Ge'ez Affixes

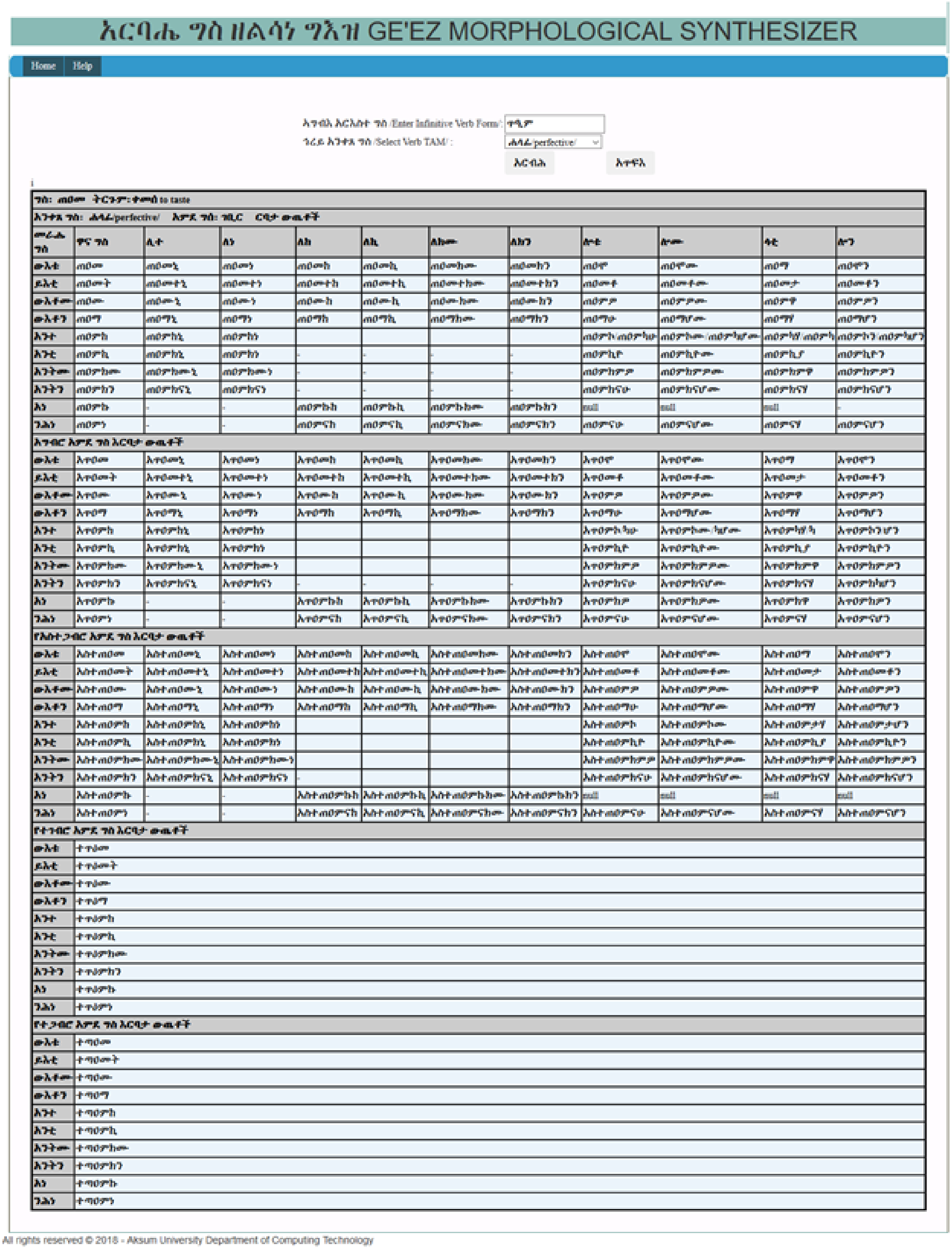

**Screenshoot of Sample Generated words from the Synthesizer**